\pdfoutput=1
\documentclass{article}
\usepackage{spconf,amsmath,graphicx}
\usepackage{threeparttable}
\usepackage{dcolumn}
\usepackage{booktabs}
\usepackage{amsmath,graphicx}
\usepackage{url}
\usepackage{cite}
\usepackage{psfrag}
\usepackage{amsfonts,amssymb}
\usepackage{amsbsy}
\usepackage{graphicx}
\usepackage{epstopdf}
\usepackage{array}
\usepackage{multirow}
\usepackage{bm}
\usepackage{subfigure}
\usepackage{verbatim}
\usepackage{colortbl}
\usepackage{stfloats}
\usepackage{algorithm}
\usepackage{algorithmic}
\usepackage{setspace}

\usepackage{bigstrut}
\usepackage{amsmath}
\usepackage{mathtools}

\def\L{{\cal L}}

\title{A Modified Perturbed Sampling Method for Local Interpretable Model-agnostic Explanation}
\name{Sheng Shi$^{\dagger}$ \qquad Xinfeng Zhang$^{\star}$ \qquad  Wei Fan$^{\dagger}$}
\address{$^{\dagger}$ AI Laboratory, Lenovo Research, Beijing 100094, China \\
$^{\star}$ University of Chinese Academy of Sciences, Beijing 100049, China}
\begin{document}
%
\maketitle
\begin{abstract}
Explainability is a gateway between Artificial Intelligence and society as the current popular deep learning models are generally weak in explaining the reasoning process and prediction results. Local Interpretable Model-agnostic Explanation (LIME) is a recent technique that explains the predictions of any classifier faithfully by learning an interpretable model locally around the prediction. However, the sampling operation in the standard implementation of LIME is defective. Perturbed samples are generated from a uniform distribution, ignoring the complicated correlation between features. This paper proposes a novel Modified Perturbed Sampling operation for LIME (MPS-LIME), which is formalized as the clique set construction problem. In image classification, MPS-LIME converts the superpixel image into an undirected graph. Various experiments show that the MPS-LIME explanation of the black-box model achieves much better performance in terms of understandability, fidelity, and efficiency.


\end{abstract}
\section{Introduction}
\label{sec:intro}
Artificial Intelligence (AI) \cite{AI1}\cite{AI2}\cite{AI3} has gone from a science-fiction dream to a critical part of our everyday life. Notably, deep learning has achieved superior performance in image classification and other perception intelligence tasks. Despite its outstanding contribution to the progress of AI, deep learning models remain mostly black boxes, which are extremely weak in explaining the reasoning process and prediction results. Nevertheless, many real-world applications are mission-critical, and users concern about how the AI solution is arriving at its decisions and insights. Therefore, model transparency and explainability are essential to ensure AI's broad adoption in various vertical domains.

There has been a recent surge in the development of explainable AI techniques\cite{One01}\cite{One02}\cite{One03}. Among them, the post hoc techniques for explaining black-box models in a human-understandable manner have received much attention in the research community\cite{local01}\cite{local02}\cite{shap}. Model-agnostic is the prominent characteristic of these methods, which generate perturbed samples of a given instance in the feature space and observe the effect of these perturbed samples on the output of the black-box classifier. In \cite{local01}, the authors proposed the Local Interpretable Model-agnostic Explanation (LIME), which explains the predictions of any classifier faithfully by fitting a linear regression model locally around the prediction. The sampling operation for LIME is a random uniform distribution, which is straightforward but defective, ignoring the correlation between features. Proper sampling operation is especially essential in natural image recognition because the visual features of natural objects exhibit a strong correlation in the spacial neighborhood, rather than a complete uniform distribution. In some cases, when most uniformly generated samples are unrealistic about the actual distribution, false information contributors lead to poorly fitting of the local explanation model.

In this paper, we propose a Modified Perturbed Sampling method for LIME (MPS-LIME), which takes into full account the correlation between features. We convert the superpixel image into an undirected graph, and then the perturbed sampling operation is formalized as the clique set construction problem. We perform various experiments on explaining Google's pre-trained Inception neural network\cite{Inception}. The experimental results show that the MPS-LIME explanation of the black-box model can achieve much better performance than LIME in terms of understandability, fidelity, and efficiency.

%

\section{MPS-LIME explanation}
\label{sec:pagestyle}

In this section, we first introduce the interpretable image representation and the modified perturbed sampling for local exploration. Then we present the explanation system of MPS-LIME.

\subsection{Interpretable Image Representation}
\label{ssec:subhead}

An interpretable representation should be understandable to observers, regardless of the underlying features used by the model. Most image classification tasks represent the image as a tensor with three color channels per pixel. Considering the poor interpretability and high computational complexity of the pixel-based representation, we adopt a superpixel based interpretable representation. Each superpixel, as the primary processing unit, is a group of connected pixels with similar colors or gray levels. Superpixel segmentation is dividing an image into some non-overlapping superpixels. More specifically, we denote $x\in{\mathbb{R}^d}$ be the original representation of an image, and binary vector $x'\in{\{{0,1}\}^{d'}}$ be its interpretable representation where $1$ indicates the presence of original superpixel and $0$ indicates an absence of original superpixel.


\subsection{A Modified Perturbed Sampling for Local Exploration}
\label{ssec:subhead}

In order to learn the local behavior of image classifier $f$, we generate a group of perturbed samples of a given instance, $x$, by activating a subset of superpixels in $x$. For the images, especially natural images, superpixel segments often correspond to the coherent regions of visual objects, showing strong correlation in a spacial neighborhood. If the activated superpixels come from an independent sampling process, we may lose much useful information to learn the local explanation models. The perturbed sampling operation in the standard implementation of LIME is to draw nonzero elements of $x'$ uniformly at random. This approach is at risk of ruining the learning process of local explanation models, since the generated samples may ignore the correlation between superpixels.

\begin{figure}
\begin{minipage}{0.32\linewidth}
  \centerline{\includegraphics[width=1.0\textwidth]{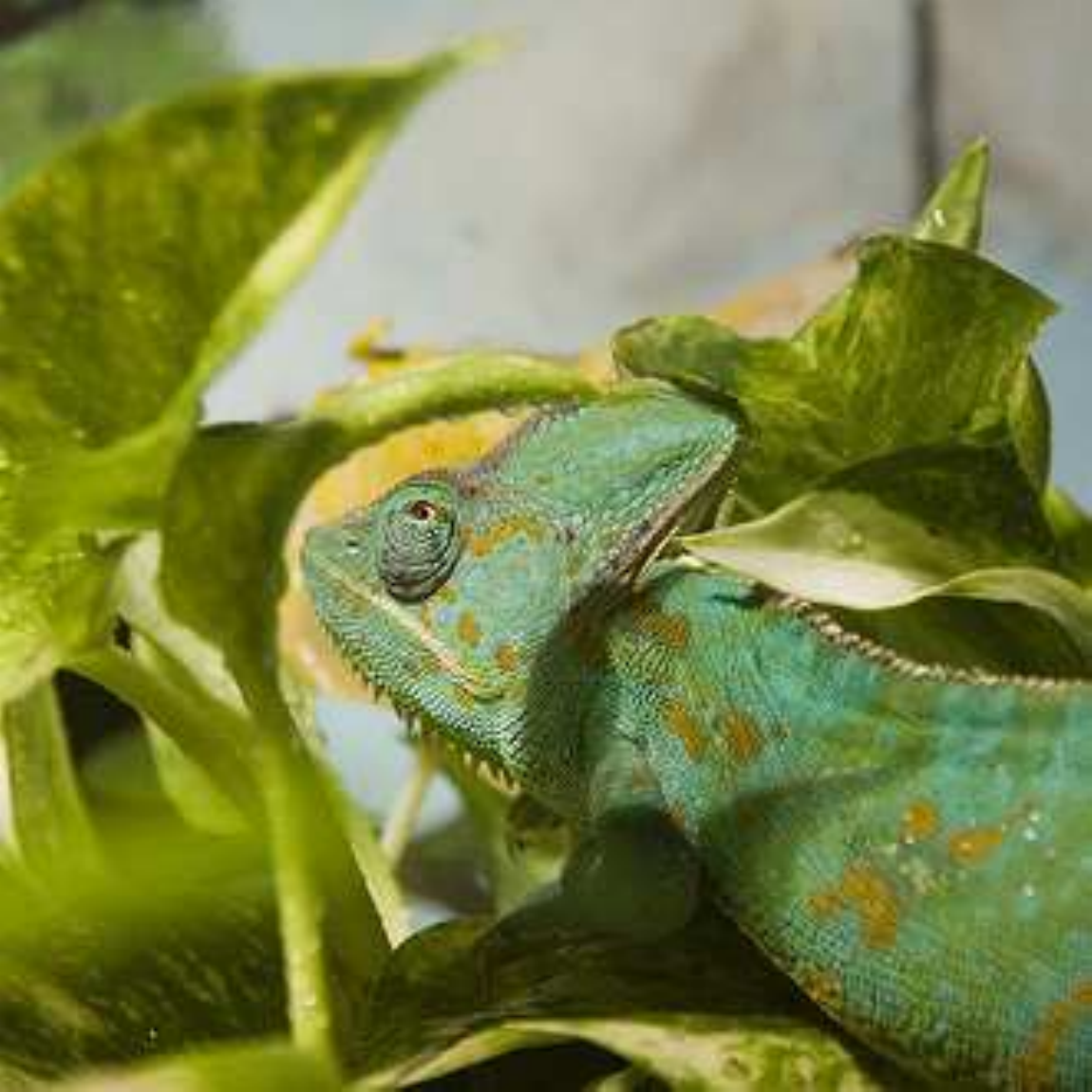}}
  \centerline{\scriptsize{(a)}}
\end{minipage}
\hfill
\begin{minipage}{0.32\linewidth}
  \centerline{\includegraphics[width=1.0\textwidth]{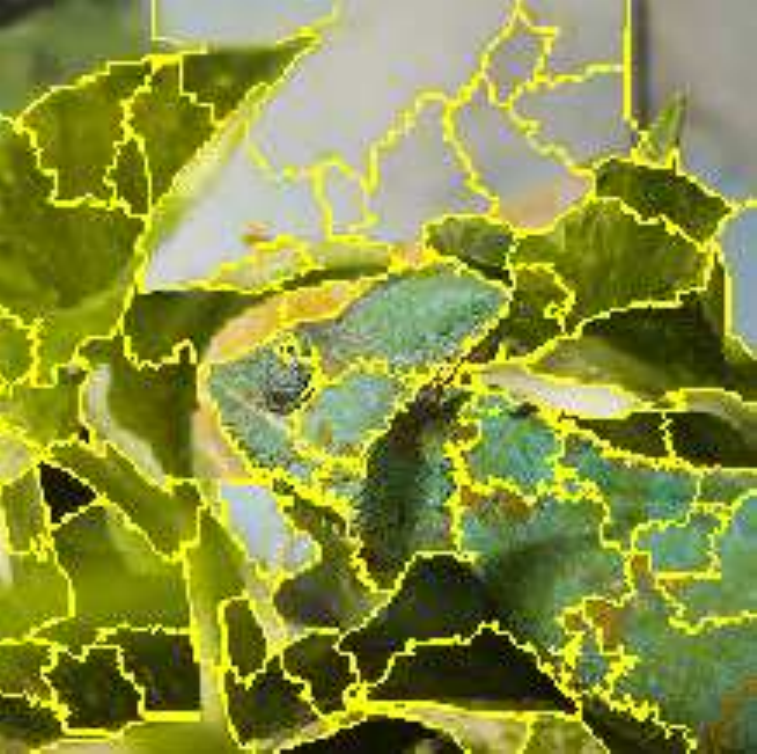}}
  \centerline{\scriptsize{(b)}}
\end{minipage}
\hfill
\begin{minipage}{0.32\linewidth}
  \centerline{\includegraphics[width=1.0\textwidth]{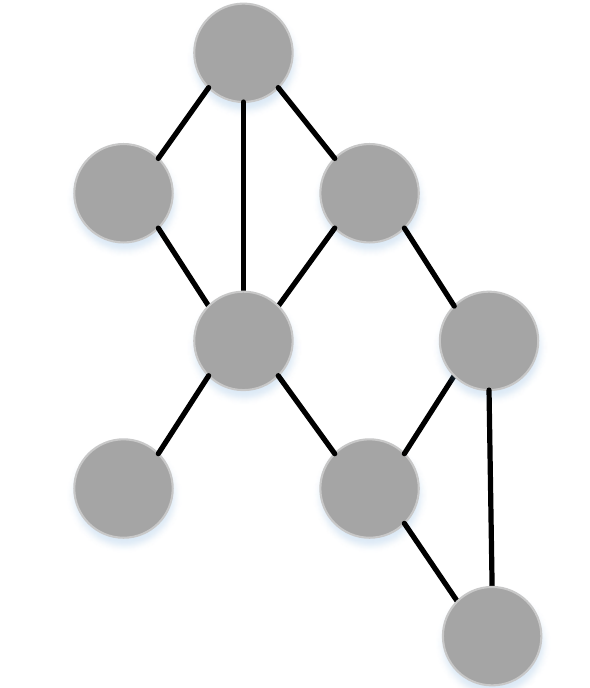}}
  \centerline{\scriptsize{(c)}}
\end{minipage}
\caption{\small{(a) Pixel-based image; (b) Superpixel image; (c) Constructing a graph of all superpixel blocks}}
\label{fig:1}
\end{figure}

In this section, we propose a modified perturbed sampling method, which takes into full account the correlation among superpixels. Firstly, we convert the superpixel segments into an undirected graph. Specifically, as shown in Figure~\ref{fig:1}, the superpixel segments are represented as vertices of a graph whose edges connect to only those adjacent segments. Considering a graph $G=(V, E)$, where $V$ and $E$ are the sets of vertices and undirected edges, with cardinalities $|V|=d'$ and $|E|$, a subset of $V$ can be represented by a binary vector $z'\in{\{{0,1}\}^{d'}}$, where $1$ indicates that vertice is in the subset.

The modified perturbed sampling operation is formalized as finding the clique $C$ ($C\subseteq V$), where every two vertices are adjacent. Since the cardinality of maximum clique of the constructed graph is $3$, the clique $C$ consists of three subset $C=C_1 \cup C_2 \cup C_3$. The three subsets are as follows: $C_1$ is the subset that only contains one vertice. $C_2$ is the subset that only contains two vertices that are connected by an edge. $C_3$ is the subset that contains three vertices, and every two vertices are adjacent (Figure 2). In this paper, we use the Depth-First Search (DFS) method to get the clique $C$. Algorithm 1 shows a simplified workflow diagram.

\begin{algorithm}[h]
\caption{\small{DFS(graph, V, v, clique, visited, start, path, n)}}
\begin{algorithmic}[1]
\STATE visited[v] $\leftarrow$ True
\IF{n==0}
   \STATE visited[v] $\leftarrow$ False
   \IF{graph[v][start]==1}
      \STATE c=copy.deepcopy(path)
      \STATE clique.append(c)
      \STATE path.pop()
      \STATE return clique
   \ELSE
      \STATE return clique
   \ENDIF
\ENDIF

\FOR{i in range(V)}
   \IF{visited[i]==False and graph[v][i]==1}
      \STATE path.append(i)
      \STATE pp=DFS(graph, V, v, clique, visited, start, path, n-1)
      \FOR {node in path}
         \IF {visited[node]==False}
            \STATE path.remove(node)
         \ENDIF
      \ENDFOR
   \ENDIF
\ENDFOR

\STATE visited[v]=False
\RETURN clique
\end{algorithmic}
\end{algorithm}

Since there is a strong correlation between the adjacent superpixel image segments, the clique $C$ set construction can take into full account the various types of neighborhood correlation. Moreover, the number of perturbed samples of MPS-LIME is much smaller than that in the current implementation of LIME, which significantly reduces the runtime.

\begin{figure}
\begin{minipage}{0.32\linewidth}
  \centerline{\includegraphics[width=1.0\textwidth]{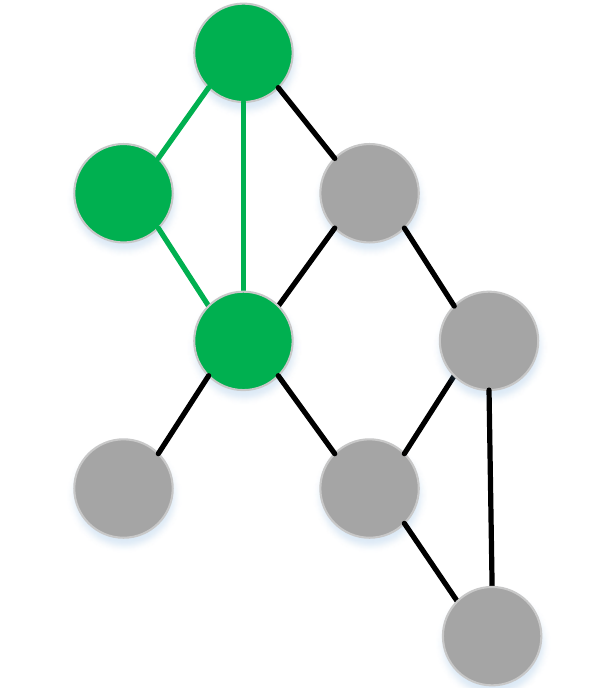}}
\end{minipage}
\hfill
\begin{minipage}{0.32\linewidth}
  \centerline{\includegraphics[width=1.0\textwidth]{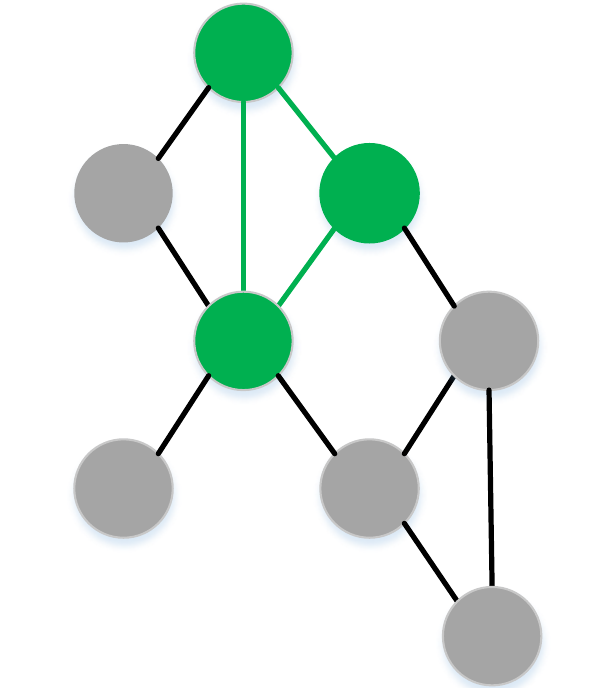}}
\end{minipage}
\hfill
\begin{minipage}{0.32\linewidth}
  \centerline{\includegraphics[width=1.0\textwidth]{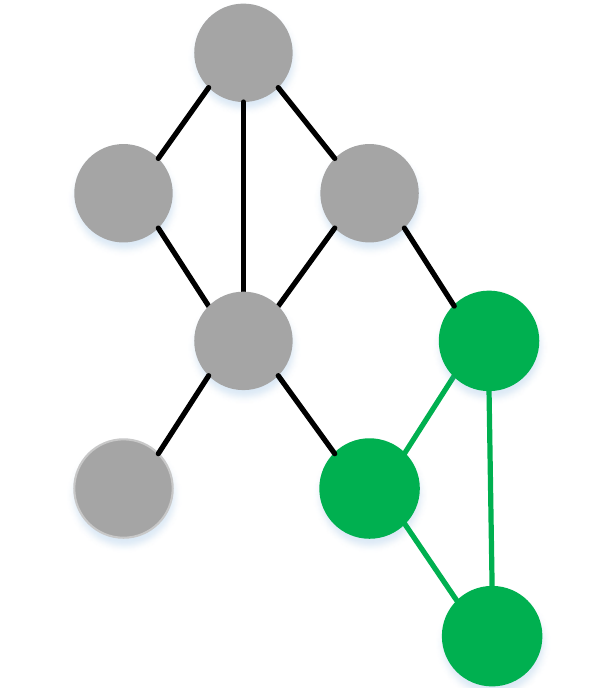}}
\end{minipage}
\caption{\small{The clique $C_3$ which is the subset that contains three vertices, where every two vertices are adjacent (marked green)}}
\label{fig:2}
\end{figure}

\subsection{Explanation System of MPS-LIME}
\label{ssec:subhead}

The goal of the explanation system is to identify an interpretable model over the interpretable representation that is locally faithful to the classifier. We denote the original image classification model being explained by $f$, and the interpretable model by $g$. This problem can be formalized as an optimization problem:
\begin{equation}
\xi(x)=argmin\quad{L(f,g,\pi_x)+ \Omega(g)},
\end{equation}
where the locality fidelity loss $L(f,g,\pi_x)$ is calculated by the locally weighted square loss:
\begin{equation}
\L(f,g,\pi_x)=\sum_{z,z'\in Z}e^{(-D(x,z)^2/{\sigma}^2)}(f(z)-g (z'))^2.\\
\end{equation}
The database $Z$ is composed of perturbed samples $z'\in{\{{0,1}\}^{d'}}$ which are sampled around $x'$ by the method described in Section 3.2. Given a perturbed sample $z'$, we recover the sample in the original representation $z\in {\mathbb{R}^d}$ and get $f(z)$. Moreover, $\pi_x(z)$ is the $L_2$ distance function to capture locality.

Algorithm 2 shows a simplified workflow diagram of MPS-LIME. Firstly, MPS-LIME gets the superpixel image by using the segment method. Then it converts the superpixel image segments into an undirected graph. The database $Z$ is constructed by finding the clique of an undirected graph, which is solved by the DFS method. Finally, MPS-LIME gets the $g$ by using the K-LASSO method, which is the same as that in LIME  \cite{local01}.

\begin{algorithm}[h]
\caption{\small{Modified Perturbed Sampling Method for Local interpretable model-agnostic explanation (MPS-LIME)}}
\label{alg::conjugateGradient}
\begin{algorithmic}[1]
\footnotesize
\REQUIRE
Classifier $f$,
Instance $x$,
Length of explanation $K$
\STATE get superpixel image $x'$ by segment method
\STATE get $f(x')$ by classifier $f$
\STATE convert the superpixel image segments into an undirected graph
\STATE initial $Z \leftarrow \{\}$
\STATE construct the clique $C$ by DFS method
\FOR {$z'\in C$}
    \STATE get $z$ by recovering $z'$
    \STATE $Z \leftarrow Z \cup (z'_i,f(z_i),\pi_x(z_i))$
\ENDFOR
\STATE get $\omega$ by $\leftarrow$ K-Lasso(Z, K)
\RETURN $\omega$
\end{algorithmic}
\end{algorithm}
\section{EXPERIMENTAL RESULTS}
\label{sec:typestyle}
In this section, we perform various experiments on explaining the predictions of Google's pre-trained Inception neural network\cite{Inception}. We compare the experimental results between LIME and MPS-LIME in terms of understandability, fidelity, and efficiency.

\subsection{Measurement criterion of interpretability}
\label{ssec:subhead}
Fidelity, understandability, and efficiency are three important goals for interpretability\cite{ref01}\cite{ref02}. An explainable model with good interpretability should be faithful to the original model, understandable to the observer, and graspable in a short time so that the end-user can make wise decisions. Mean Absolute Error (MAE) and Coefficient of determination $R^2$ are two import measures of fidelity. MAE is the absolute error between the predicted value and true value, which can reflect the predictive accuracy well,
\begin{equation}
MAE=\frac{1}{n}\sum_{i=1}^n|y_{true(i)}-y_{pred(i)}|.
\end{equation}
$R^2$ is calculated by Total Sum of Squares (SST) and Error Sum of Squares (SSE):
\begin{eqnarray}
\nonumber & R^2=1-SSE/SST\\
\nonumber& SSE=\sum_{i=1}^n(y_{true(i)}-y_{pred(i)})^2\\
& SST=\sum_{i=1}^n(y_{true(i)}-y_{mean(i)})^2,
\end{eqnarray}
where $y_{true}$ is the true value, $y_{pred}$ is the predicted value and $y_{mean}$ is the mean value of true value. The best $R^2$ is $1.0$. The closer the score is to $1.0$, the better the performance of fidelity is to explainer.

\subsection{Google's Inception neural network on Image-net database}
\label{ssec:subhead}

We explain image classification predictions made by Google's pre-trained Inception neural network\cite{Inception}. The first row in Figure~\ref{fig:2} shows six original images. The second row and third row are the superpixels explanations by LIME and MPS-LIME, respectively. The explanations highlight the top $5$ superpixel segments, which have the most considerable positive weights towards the predictions (K=5).

Table~\ref{tab:1} lists the MAE of LIME and MPS-LIME. We find some of the predictive probability values of LIME is bigger than $1.0$.  This is because LIME adopts a sparse linear model to fit the perturbed samples, and has no more constraints such as the probability values distribution should range between 0 and 1. Comparing to LIME, we can see that MPS-LIME provides better predictive accuracy than LIME. Besides, $R^2$ of LIME and MPS-LIME are listed in Table~\ref{tab:1}. The closer the score is to $1.0$, the better the performance of fidelity is to an explainer. The $R^2$ of MPS-LIME is around $0.9$, which is much bigger than LIME. By comparing the MAE and $R^2$ of two algorithms, we can conclude that MPS-LIME has better fidelity than LIME.

Efficiency is highly related to the time necessary for a user to grasp the explanation. The runtime of LIME and MPS-LIME are shown in Table~\ref{tab:2}, which shows that the runtime of MPS-LIME is nearly half as the runtime of LIME. We can conclude from the above results that MPS-LIME not only has a higher fidelity but also take less time than LIME.

\begin{figure*}
\begin{minipage}{0.162\linewidth}
  \centerline{\includegraphics[width=1.0\textwidth]{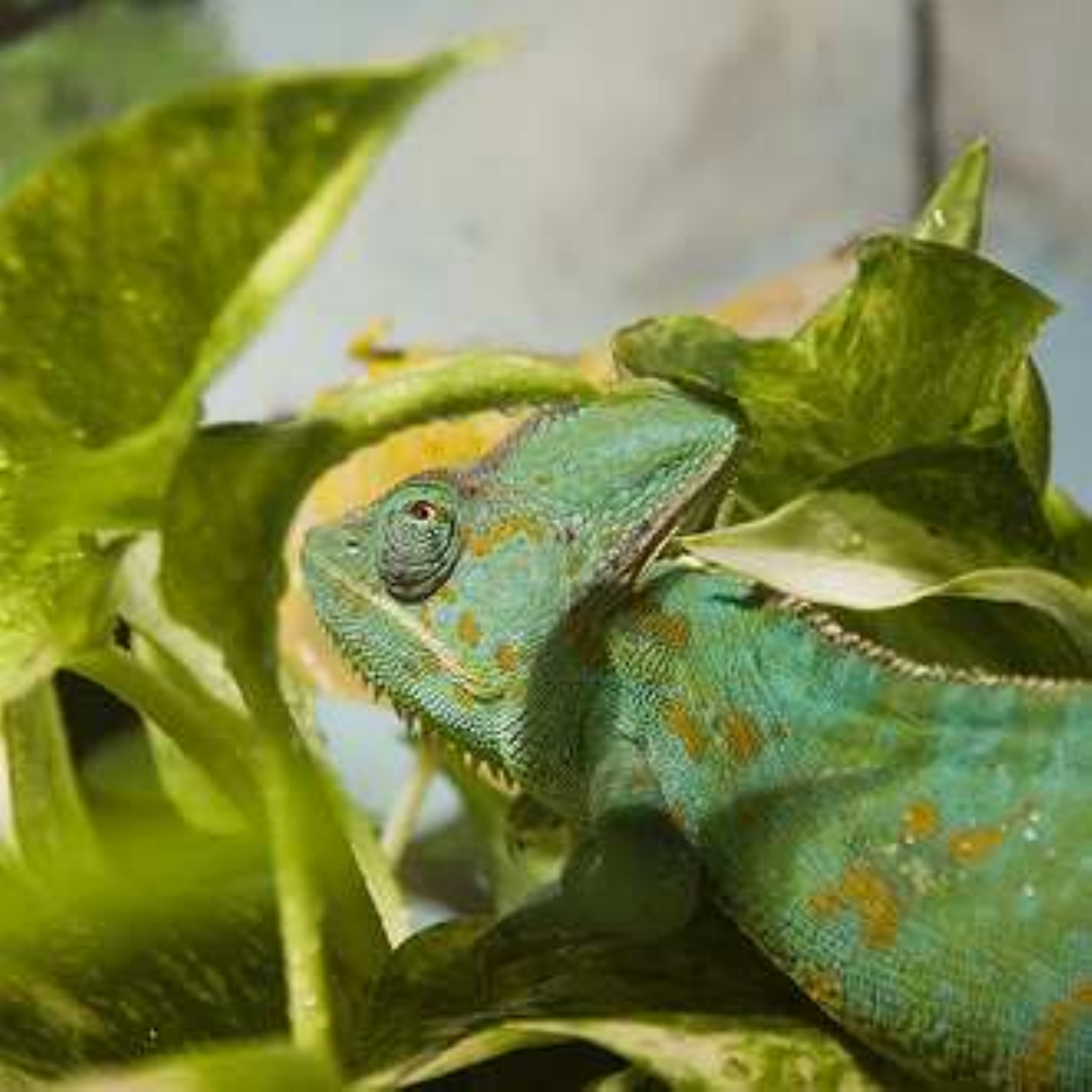}}
\end{minipage}
\hfill
\begin{minipage}{0.162\linewidth}
  \centerline{\includegraphics[width=1.0\textwidth]{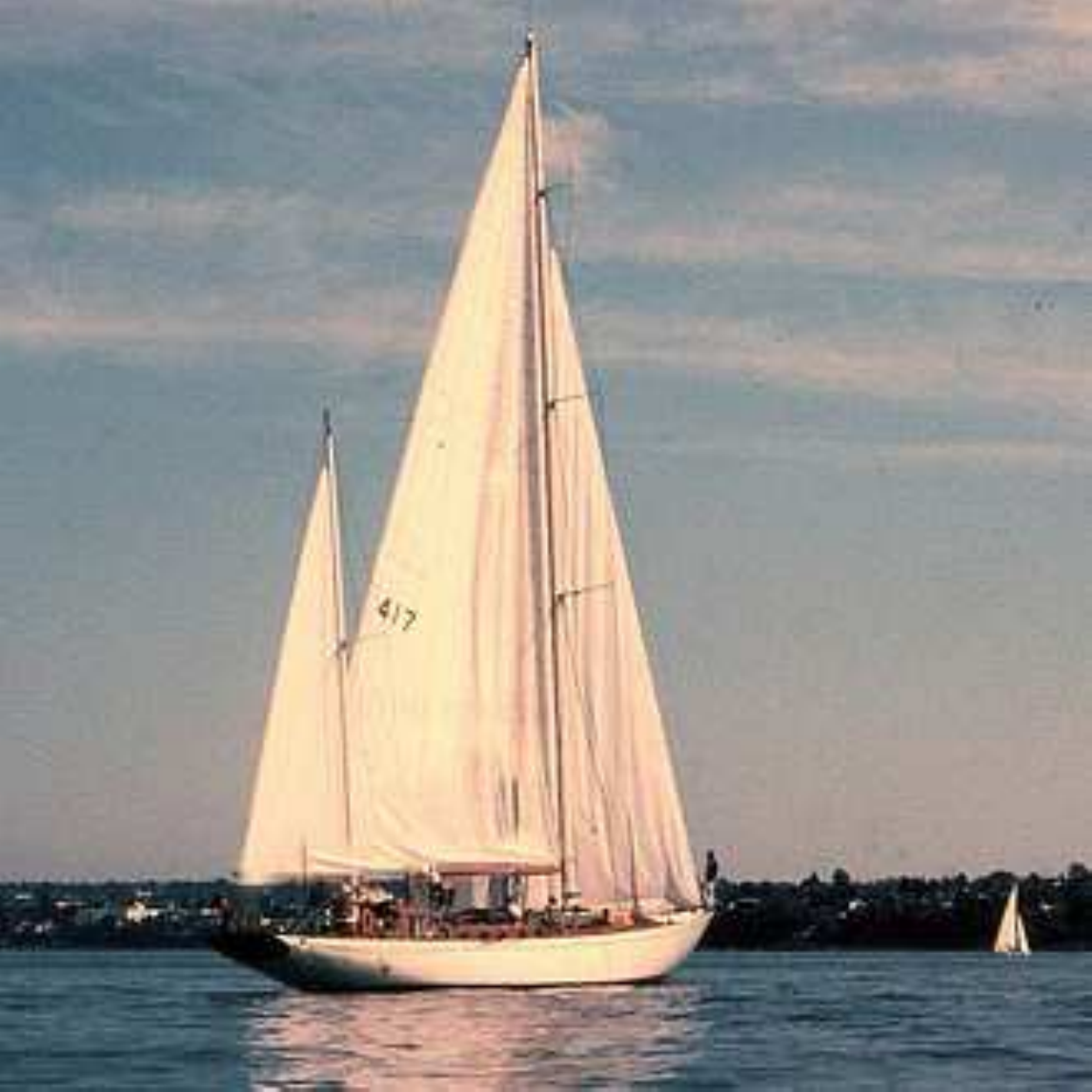}}
\end{minipage}
\hfill
\begin{minipage}{0.162\linewidth}
  \centerline{\includegraphics[width=1.0\textwidth]{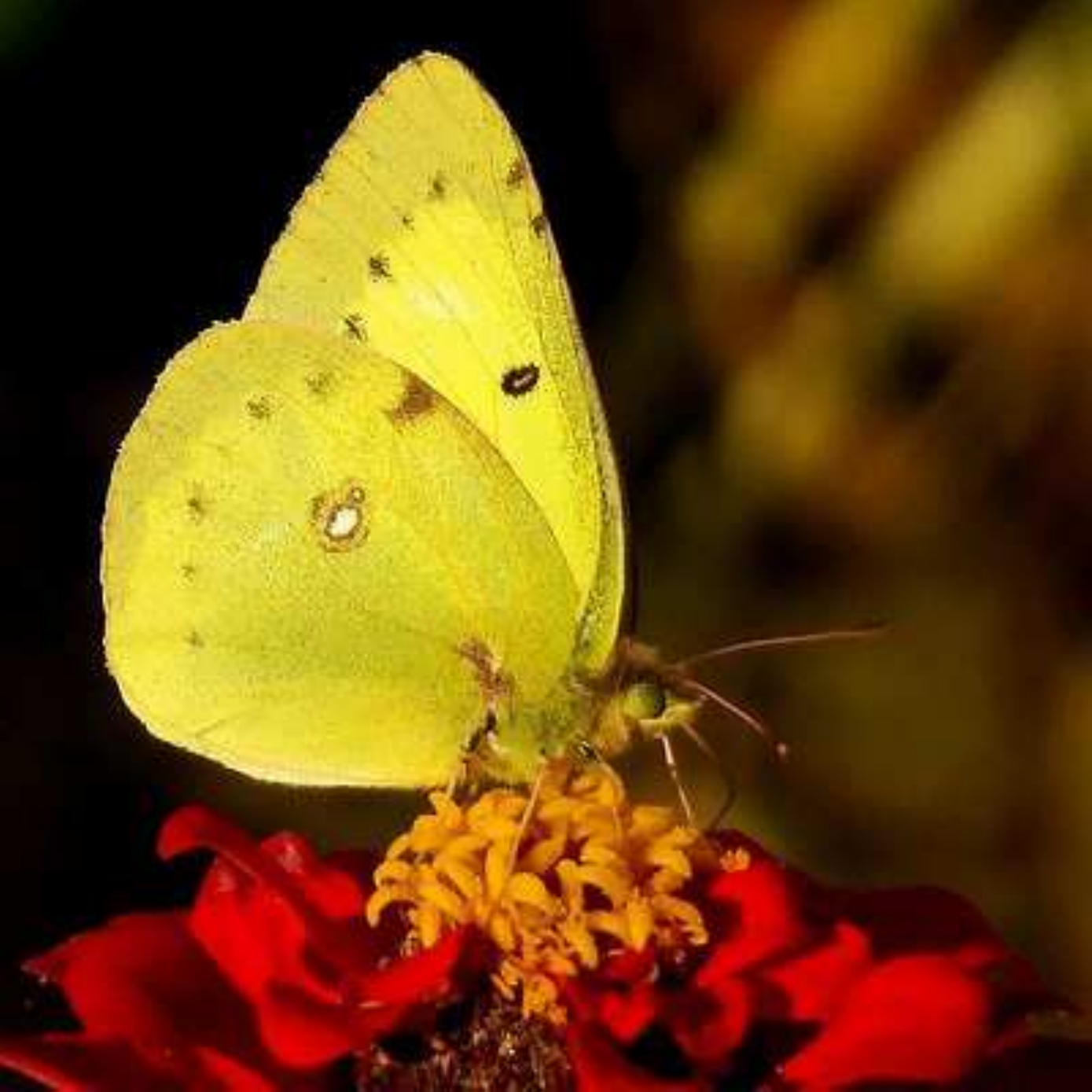}}
\end{minipage}
\hfill
\begin{minipage}{0.162\linewidth}
  \centerline{\includegraphics[width=1.0\textwidth]{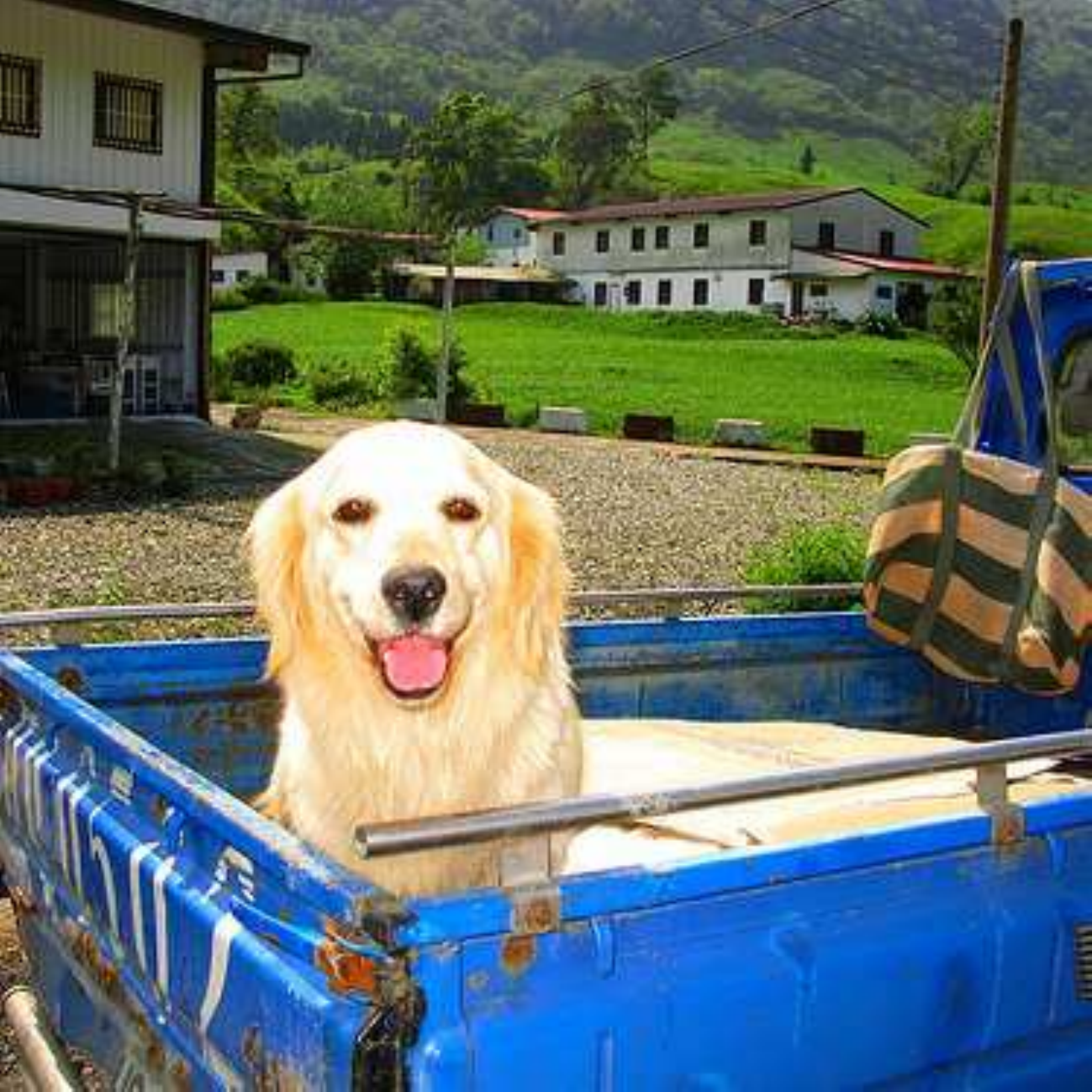}}
\end{minipage}
\hfill
\begin{minipage}{0.162\linewidth}
  \centerline{\includegraphics[width=1.0\textwidth]{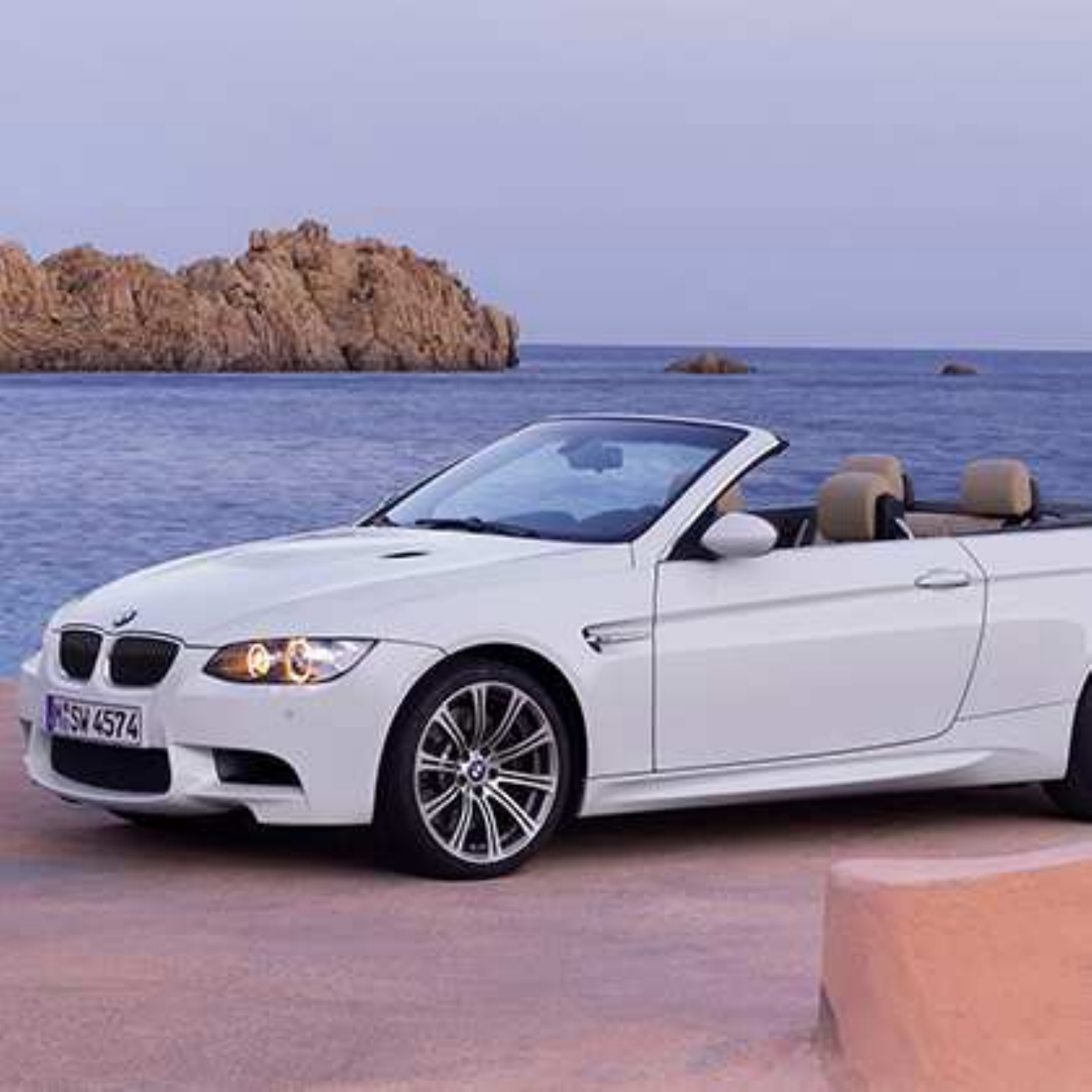}}
\end{minipage}
\hfill
\begin{minipage}{0.162\linewidth}
  \centerline{\includegraphics[width=1.0\textwidth]{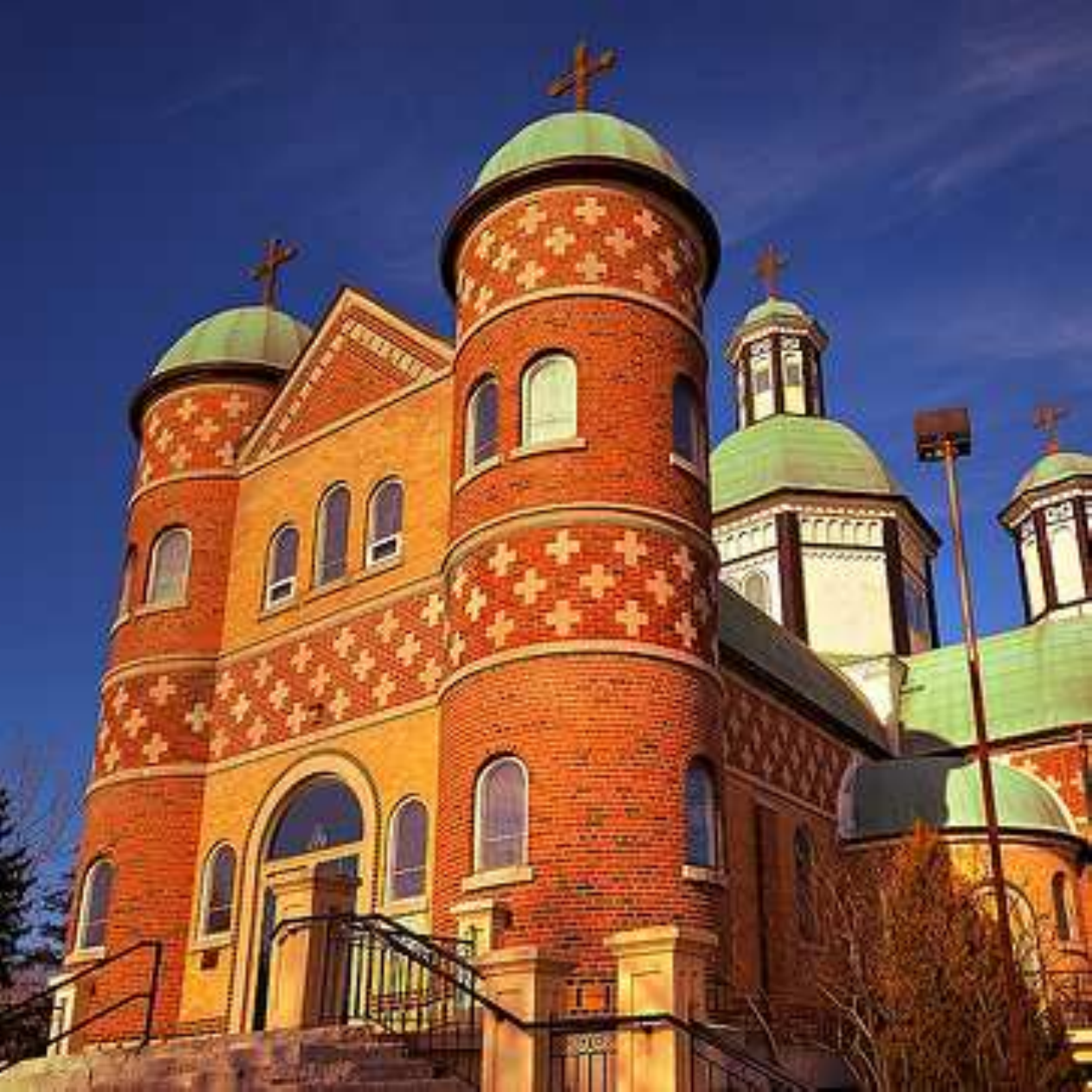}}
\end{minipage}
\\
\vfill
\begin{minipage}{0.162\linewidth}
  \centerline{\includegraphics[width=1.0\textwidth]{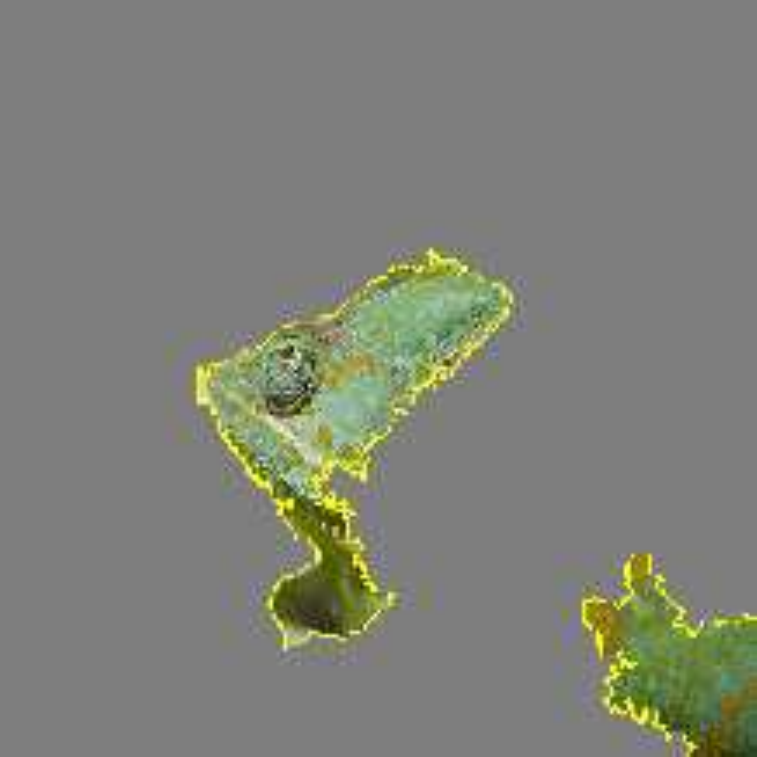}}
\end{minipage}
\hfill
\begin{minipage}{0.162\linewidth}
  \centerline{\includegraphics[width=1.0\textwidth]{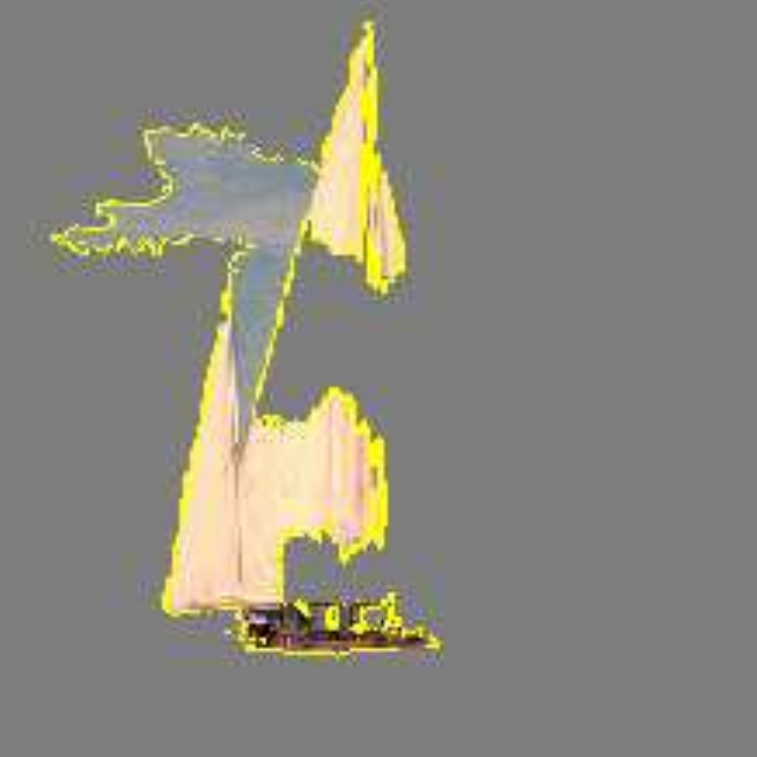}}
\end{minipage}
\hfill
\begin{minipage}{0.162\linewidth}
  \centerline{\includegraphics[width=1.0\textwidth]{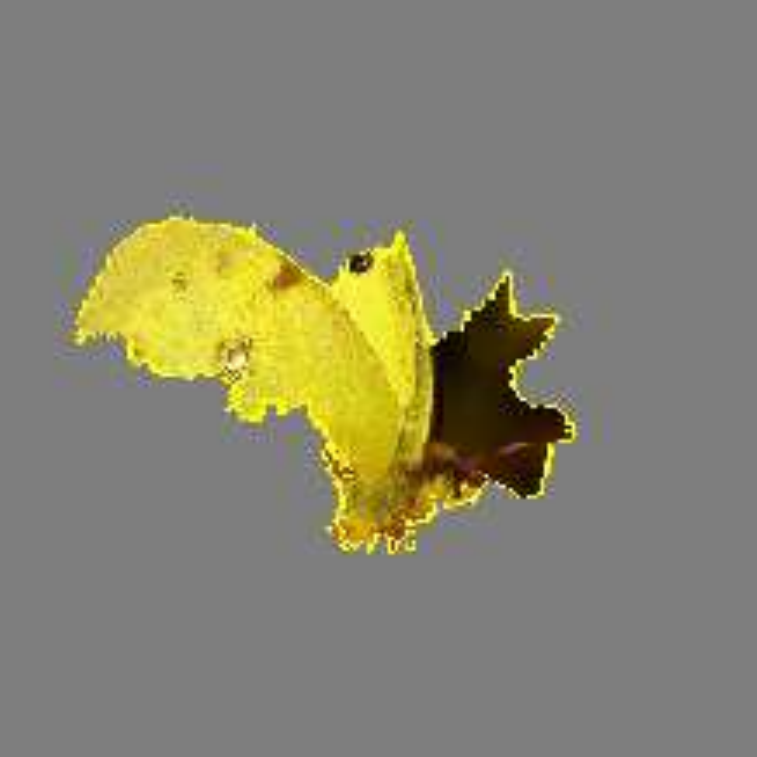}}
\end{minipage}
\hfill
\begin{minipage}{0.162\linewidth}
  \centerline{\includegraphics[width=1.0\textwidth]{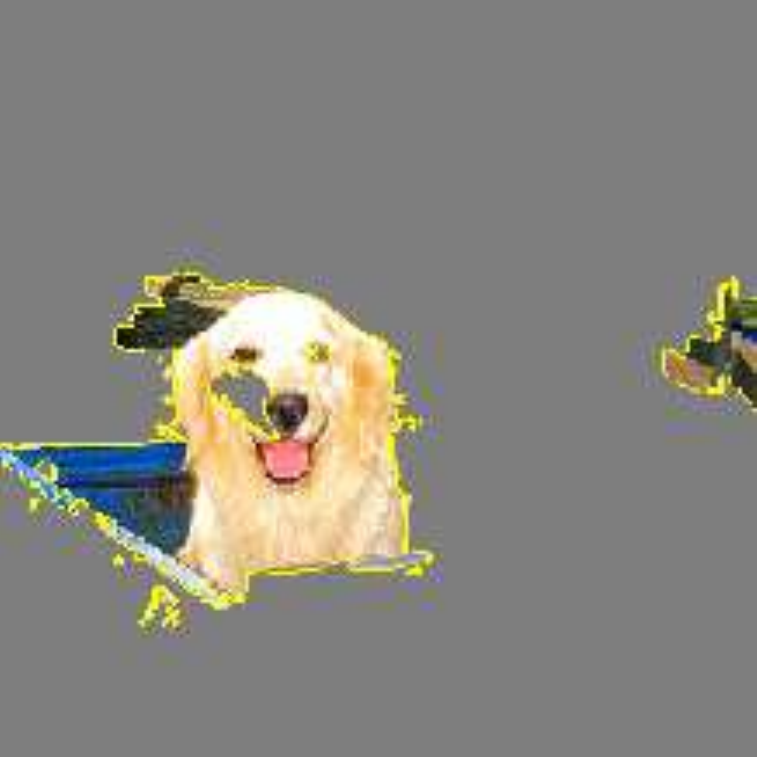}}
\end{minipage}
\hfill
\begin{minipage}{0.162\linewidth}
  \centerline{\includegraphics[width=1.0\textwidth]{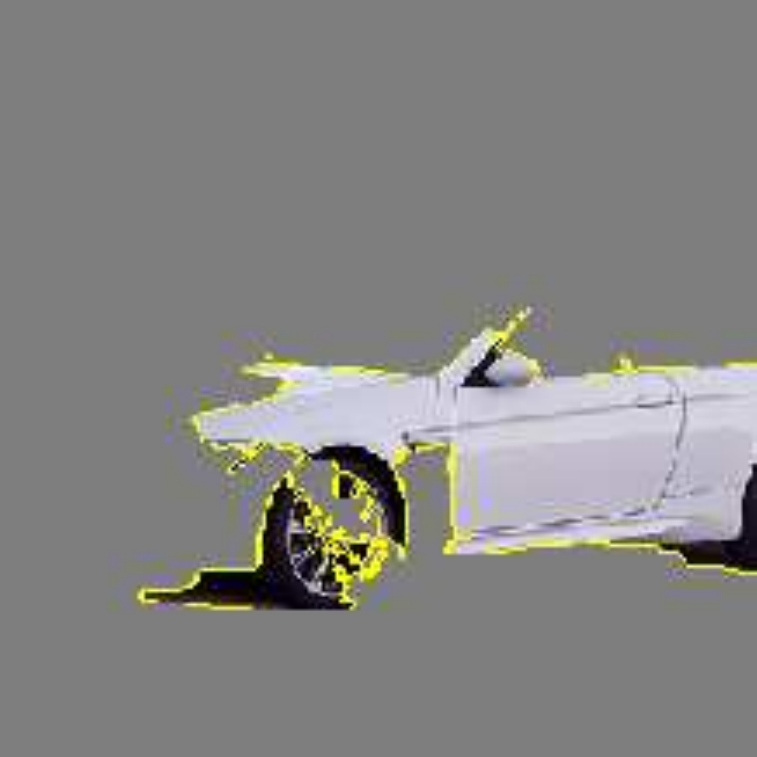}}
\end{minipage}
\hfill
\begin{minipage}{0.162\linewidth}
  \centerline{\includegraphics[width=1.0\textwidth]{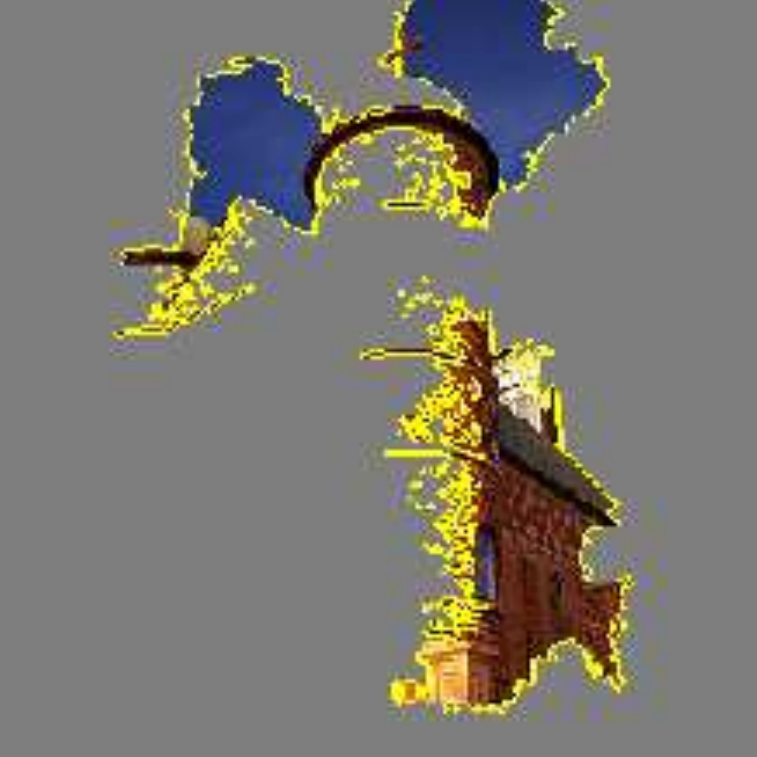}}
\end{minipage}
\\
\vfill
\begin{minipage}{0.162\linewidth}
  \centerline{\includegraphics[width=1.0\textwidth]{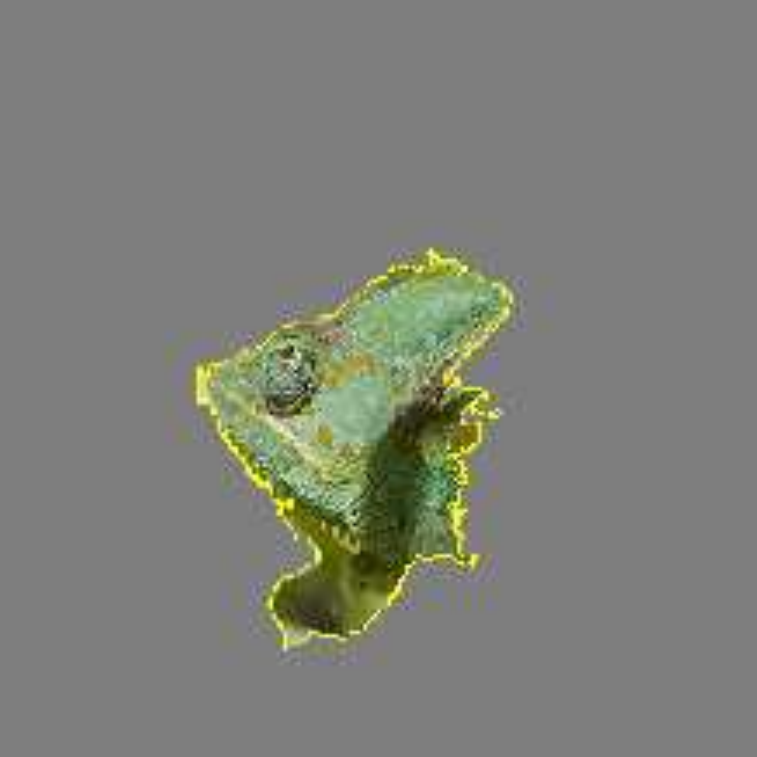}}
\end{minipage}
\hfill
\begin{minipage}{0.162\linewidth}
  \centerline{\includegraphics[width=1.0\textwidth]{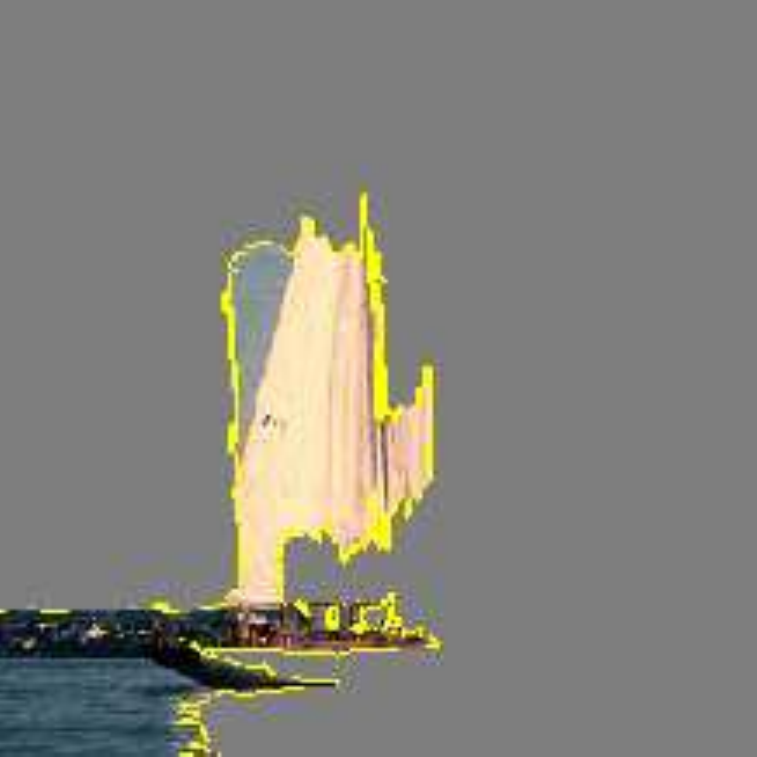}}
\end{minipage}
\hfill
\begin{minipage}{0.162\linewidth}
  \centerline{\includegraphics[width=1.0\textwidth]{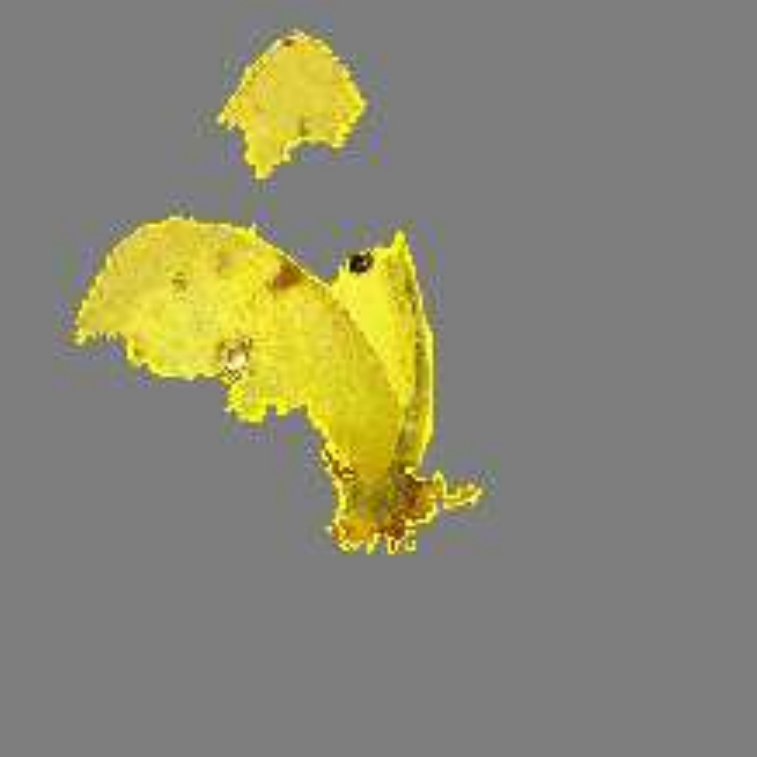}}
\end{minipage}
\hfill
\begin{minipage}{0.162\linewidth}
  \centerline{\includegraphics[width=1.0\textwidth]{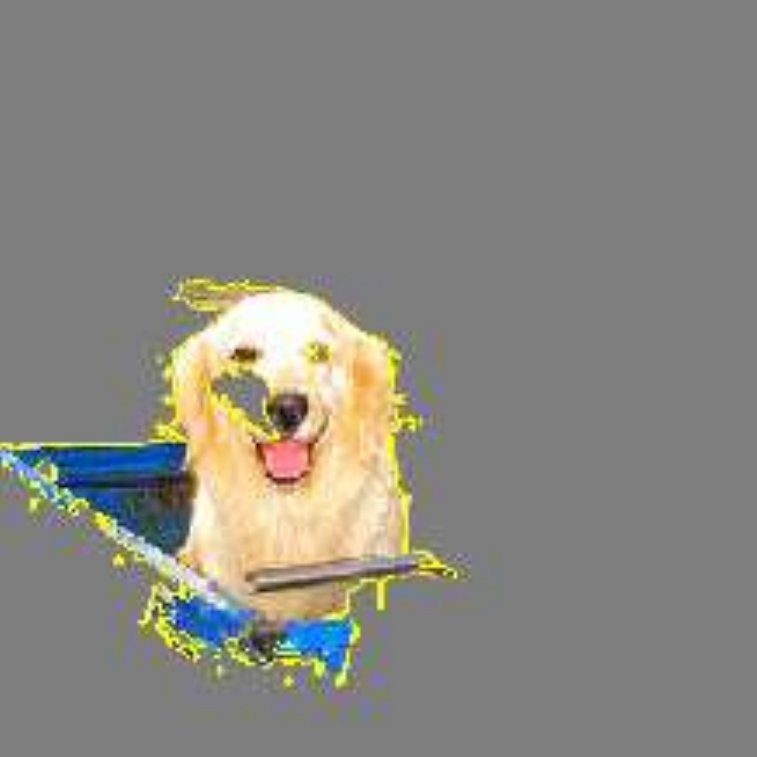}}
\end{minipage}
\hfill
\begin{minipage}{0.162\linewidth}
  \centerline{\includegraphics[width=1.0\textwidth]{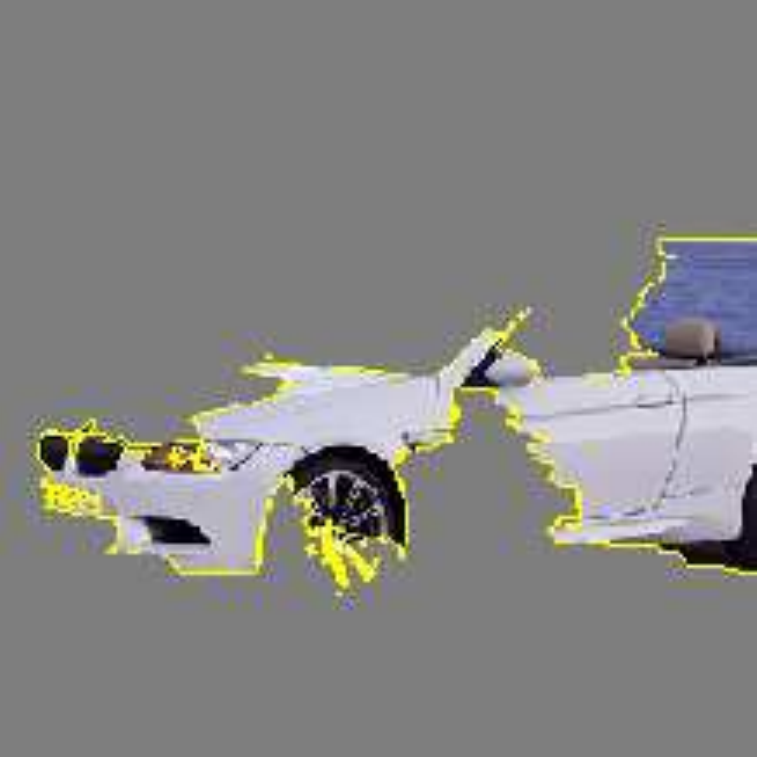}}
\end{minipage}
\hfill
\begin{minipage}{0.162\linewidth}
  \centerline{\includegraphics[width=1.0\textwidth]{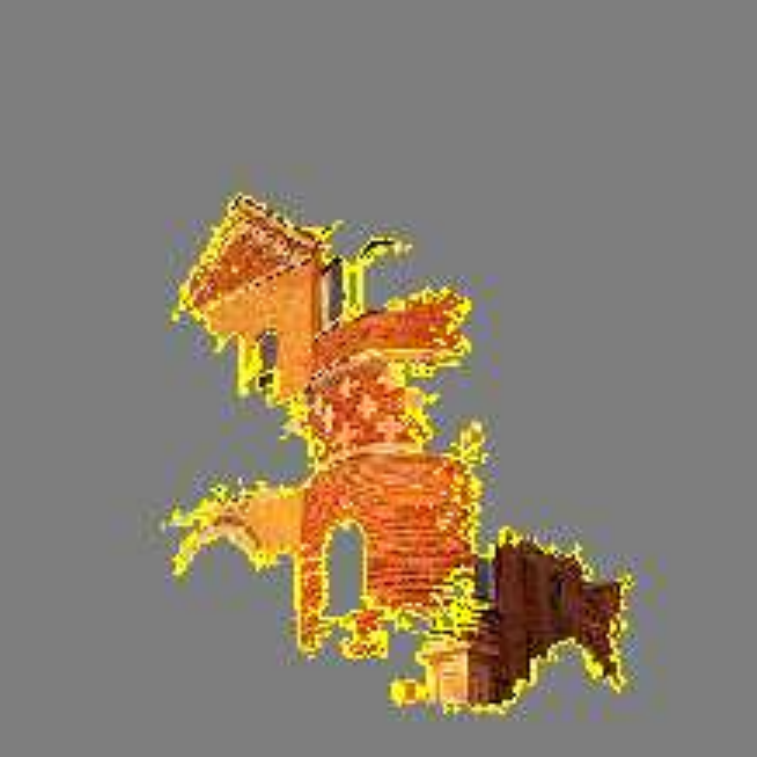}}
\end{minipage}
\caption{\small{Explaining image classification predictions made by Google's Inception neural network. The first row shows $6$ original images, and the top $1$ class predicted of the original images are $African$ $chameleon (p=0.9935)$, $yawl (p=0.6077)$, $sulphur$ $butterfly (p=0.9431)$, $golden$ $retriever (p=0.5641)$, $convertible (p=0.9356)$, $castle (p=0.7646)$. The second row shows the superpixels explanations by LIME (K=5). The third row shows the superpixels explanations by MPS-LIME (K=5).}}
\label{fig:2}
\end{figure*}

\begin{table}
  \centering\footnotesize
  \caption{\small{The MAE of LIME and MPS-LIME on Google's pre-trained Inception neural network}}
    \begin{tabular}{|c|c|c|c|c|}
    \hline
     & true prob (Inception) & pred prob  & MAE & $R^2$ \bigstrut\\
    \hline
    LIME &\multirow{2}{*}{$p_{chameleon}=0.9935$}&1.6285 &0.6350 &0.6885  \bigstrut\\
    MPS-LIME  & & 0.9783 & 0.0152 &0.8944 \bigstrut\\
    \hline
    LIME &\multirow{2}{*}{$p_{yawl}=0.6077$}& 0.8291 & 0.2214 &0.4531 \bigstrut\\
    MPS-LIME  & & 0.5973 & 0.0104 &0.9825\bigstrut\\
    \hline
    LIME &\multirow{2}{*}{$p_{butterfly}=0.9431$}& 1.6668 & 0.7237 &0.636 \bigstrut\\
    MPS-LIME  & & 0.9284 & 0.0147 &0.9222\bigstrut\\
    \hline
    LIME &\multirow{2}{*}{$p_{retriever}=0.5641$}& 0.4822 &0.0819 &0.6958 \bigstrut\\
    MPS-LIME  & & 0.5568 & 0.0073 &0.9304\bigstrut\\
    \hline
    LIME &\multirow{2}{*}{$p_{convertible}=0.9356$}& 1.2854 & 0.3498 &0.7407 \bigstrut\\
    MPS-LIME  & & 0.9203 & 0.0153 &0.9925\bigstrut\\
    \hline
    LIME &\multirow{2}{*}{$p_{castle}=0.7646$}& 1.0166 & 0.2520  &0.3535 \bigstrut\\
    MPS-LIME  & & 0.7531 & 0.0115 &0.9155\bigstrut\\
    \hline
    \end{tabular}
  \label{tab:1}
\end{table}


\begin{table}
  \centering\footnotesize
  \caption{\small{The runtime of LIME and MPS-LIME on Google's pre-trained Inception neural network}}
    \begin{tabular}{|c|c|c|c|c|c|c|}
    \hline
     & img1 & img2  & img3  & img4 &img5 &img6 \bigstrut\\
    \hline
    LIME &232.20 &230.45 &245.36 &264.51 &223.79  &226.58\bigstrut\\
    \hline
    MPS-LIME &91.02 & 113.85 &109.29 & 154.57 &117.21 &152.84 \bigstrut\\
    \hline
    \end{tabular}
  \label{tab:2}
\end{table}

\section{Conclusion and Future Work}
\label{sec:majhead}

The sampling operation for local exploration in the current implementation of LIME is a random uniform sampling, which possibly generates unrealistic samples ruining the learning of local explanation models. In this paper, we propose a modified perturbed sampling method MPS-LIME, which takes into full account the correlation between features. We convert the superpixel image into an undirected graph, and then the perturbed sampling operation is formalized as the clique set construction problem. We perform various experiments on explaining the random-forest classifier and Google's pre-trained Inception neural network. Various experiment results show that the MPS-LIME explanation of multiple black-box models can achieve much better performance in terms of understandability, fidelity, and efficiency.

There are some avenues of future work that we would like to explore. This paper only describes the modified perturbed sampling method for image classification. We will apply the similar idea to text processing and structural data analytics. Besides, we will improve other post hoc explanations techniques that rely on input perturbations such as SHAP and propose a general optimization scheme.


\newpage
\bibliographystyle{IEEEbib}
\bibliography{strings,refs}

\end{document}